\documentclass[conference]{IEEEtran}
\usepackage{cite}
\usepackage{amsmath,amssymb,amsfonts}
\usepackage{algorithmic}
\usepackage{algorithm}
\usepackage{graphicx}
\usepackage{textcomp}
\usepackage{xcolor}
\usepackage{url}

\usepackage{flushend}

\usepackage[font=footnotesize,labelfont=bf]{subcaption}

\begin{document}
\title{
Zero-Shot Vehicle Model Recognition via Text-Based Retrieval-Augmented Generation
}
\date{}

\author{
Wei-Chia Chang\IEEEauthorrefmark{1} and Yan-Ann Chen\IEEEauthorrefmark{1}\\
\IEEEauthorrefmark{1} Yuan Ze University, Taiwan \\
Email: s1113341@mail.yzu.edu.tw, chenya@saturn.yzu.edu.tw}
\maketitle

\begin{abstract}
Vehicle make and model recognition (VMMR) is an important task in intelligent transportation systems, but existing approaches struggle to adapt to newly released models. Contrastive Language–Image Pretraining (CLIP) provides strong visual–text alignment, yet its fixed pretrained weights limit performance without costly image-specific finetuning. We propose a pipeline that integrates vision–language models (VLMs) with Retrieval-Augmented Generation (RAG) to support zero-shot recognition through text-based reasoning. A VLM converts vehicle images into descriptive attributes, which are compared against a database of textual features. Relevant entries are retrieved and combined with the description to form a prompt, and a language model (LM) infers the make and model. This design avoids large-scale retraining and enables rapid updates by adding textual descriptions of new vehicles. Experiments show that the proposed method improves recognition by nearly 20\% over the CLIP baseline, demonstrating the potential of RAG-enhanced LM reasoning for scalable VMMR in smart-city applications.
\end{abstract}

\section{Introduction}
In recent years, the rapid growth of artificial intelligence has accelerated the development of smart cities. Among various applications, accurate recognition of vehicle makes and models is useful for intelligent transportation systems. Timely identification assists in traffic accident analysis, improves traffic-flow management, and supports law enforcement when license plates are missing or deliberately altered. Despite these benefits, vehicle recognition remains challenging due to the continuous introduction of new models and the variability of visual appearances across different conditions.

Previous studies on vehicle recognition relied on computer-vision classifiers trained on handcrafted or deep visual features \cite{CV1-Traffic, CV2-BIGNet, CV-3-ResNet}. These approaches required frequent retraining and lacked robustness to unseen models. More recently, multimodal image–text methods have shown promise for visual recognition by aligning visual content with linguistic descriptions. Representative frameworks such as CLIP \cite{clip} achieve strong general-purpose alignment through large-scale pretraining. However, their performance is constrained by fixed pretrained data and weights, which limits recognition of newly released vehicle models. Furthermore, reliance on global image cues reduces effectiveness when only partial features (e.g., headlights or grilles) are available, a situation common in accident or surveillance scenarios.

To address these limitations, we propose a zero-shot framework that integrates vision–language models (VLMs) with retrieval-augmented generation (RAG). A VLM converts vehicle visual attributes into textual descriptions, which are compared against a vector database of exterior features. A language model then reasons over the retrieved entries to infer the make and model, eliminating the need for additional image retraining. RAG \cite{rag} grounds the reasoning process in external knowledge, reducing hallucinations and improving accuracy. Since new vehicle models can be incorporated by simply updating textual descriptions, the framework remains lightweight, scalable, and practical for intelligent transportation systems.



The main contributions of this paper are summarized as follows:
\begin{itemize}
\item \textbf{RAG-based framework for car model recognition:} We implement a text-centric architecture that converts visual features into semantic descriptions, performs retrieval, and reasons about make and model. To the best of our knowledge, this is the first work to apply RAG methodology to vehicle identification tasks, validating the feasibility of this approach for automotive recognition.
\item \textbf{Zero-shot new car model recognition:} The proposed method can identify car models from textual descriptions of frontal features, without requiring images.
\item \textbf{Lightweight and deployable:} By using small-scale open-source models on standard consumer GPUs, the framework avoids reliance on high-performance hardware, simplifies deployment, and supports edge applications.
\end{itemize}

The remainder of this paper is organized as follows. Section~\ref{sec:method} presents the proposed pipeline design based on vision-language models and retrieval-augmented reasoning. Section~\ref{sec:results} reports the experimental setup and results, including comparisons with the CLIP baseline. Section~\ref{sec:conclude} concludes the paper and discusses potential directions for future work. The objective of this work is to demonstrate the feasibility of text-based retrieval and reasoning for zero-shot vehicle make--model recognition in intelligent transportation systems.



\section{Method Design}
\label{sec:method}
\subsection{Problem Formulation}

We formulate vehicle make and model recognition (VMMR) as a zero-shot classification task.  
Let $\mathcal{I}$ denote the domain of vehicle images and $\mathcal{Y} = \{y_1, y_2, \dots, y_K\}$ the set of target categories,  
where each $y$ encodes both the make and the model.  
For an unseen input image $x \in \mathcal{I}$, the goal is to predict the most probable label:  

\begin{equation}
y^* = \arg\max_{y \in \mathcal{Y}} F(x, y),
\end{equation}

where $F(x, y)$ denotes a scoring function that measures the semantic consistency between $x$ and a candidate label $y$.  

To approximate $F$, a vision--language encoder $E_v : \mathcal{I} \rightarrow \mathcal{T}$ converts the input $x$ into a textual description $\hat{t} \in \mathcal{T}$.  
Within the RAG framework, $\hat{t}$ is compared against a database of textual entries  
$\mathcal{D} = \{t_j\}_{j=1}^{N}$ using a similarity function $S(\cdot,\cdot)$ to retrieve the top-$k$ candidates:  

\begin{equation}
\mathcal{R}_k(\hat{t}) = \operatorname{TopK}\!\left\{ S(\hat{t}, t_j) \;\mid\; t_j \in \mathcal{D} \right\}.
\end{equation}

The retrieved entries are concatenated with $\hat{t}$ to form a prompt  
$P = [\hat{t}; \mathcal{R}_k]$, which is passed to a language model (LM) for reasoning:  

\begin{equation}
\hat{y} = \mathrm{LM}_{\mathrm{reason}}(P).
\end{equation}

Here, $\hat{y}$ serves as a practical approximation to the optimal prediction $y^*$.

Performance on a held-out test set $\{(x_i, y_i)\}_{i=1}^{N_{\mathrm{test}}}$  
is evaluated using standard classification metrics:  
\begin{equation}
\begin{aligned}
\mathrm{Accuracy} &= \frac{1}{N_{\mathrm{test}}}\sum_{i=1}^{N_{\mathrm{test}}}\mathbf{1}\{\hat{y}_i = y_i\}, \\
\mathrm{Precision} &= \frac{\mathrm{TP}}{\mathrm{TP}+\mathrm{FP}}, \qquad
\mathrm{Recall} = \frac{\mathrm{TP}}{\mathrm{TP}+\mathrm{FN}},
\end{aligned}
\end{equation}
where $\mathrm{TP}$, $\mathrm{FP}$, and $\mathrm{FN}$ denote true positives,  
false positives, and false negatives, respectively.  

This formulation defines the zero-shot classification objective,  
the retrieval-augmented reasoning process, and the evaluation metrics,  
providing a concise mathematical abstraction of the proposed pipeline.



\subsection{Pipeline Design}

The proposed pipeline performs vehicle make and model recognition through three main stages by combining vision–language understanding with RAG. The overall idea is to shift the recognition problem from the image domain into the text domain, where new vehicle models can be flexibly described and incorporated without retraining on large-scale image datasets.  

\textbf{Stage~1: Visual-to-Text Conversion.}  
A vision–language encoder first converts an input vehicle image into a natural language description that highlights its exterior attributes, such as body type, headlights, grille design, or wheel shape. By translating visual input into descriptive text, the system avoids reliance on rigid image embeddings and instead enables zero-shot generalization, since unseen models can still be characterized through descriptive language.  

\textbf{Stage~2: Retrieval of Contextual Knowledge.}  
The generated description is then compared against a textual database, which contains curated entries representing a wide variety of vehicle makes and models. Each entry encodes distinctive attributes that help discriminate between similar categories. A similarity function ranks these entries, and the top-$k$ most relevant candidates are retrieved. Because the database can be dynamically updated with textual descriptions of newly released vehicles, the system adapts rapidly to changes in the real world without retraining on additional images.  

\textbf{Stage~3: Language Model Reasoning.}  
The retrieved entries are concatenated with the generated description to form a structured prompt, which is passed to a language model (LM). The LM reasons over both the direct description of the query image and the retrieved contextual knowledge, producing the most likely make and model. This design not only enables zero-shot recognition of previously unseen vehicles but also mitigates hallucination by grounding the LM in retrieved evidence.  

In summary, the pipeline avoids image-specific fine-tuning, leverages textual retrieval for knowledge enrichment, and ensures that recognition remains scalable, lightweight, and adaptable to the continual introduction of new vehicle models in intelligent transportation systems.  



\section{Experimental Results \label{sec:results}}

\subsection{Dataset Introduction}
Due to the fact that existing car datasets, such as CompCars \cite{compcars}, mostly contain models that have been on the market for some time, they are not suitable for evaluating the ability to recognize new car models. To address this, we collected 10 recently released car models from the web, with specific selection criteria to ensure these were entirely new models launched after 2023, rather than facelifts or derivatives of existing models. This selection strategy ensures that the CLIP model has no prior training exposure to these vehicles, providing a true zero-shot evaluation scenario. For each model, we selected 10 images for the test set, resulting in a total of 100 test images. Additionally, one image per model was used to form the training set, which serves to generate visual feature descriptions and build the retrieval database. The car models used in this study are listed in Table \ref{tabCMM}.
\begin{table}[!t]
\centering
\caption{THE CAR MAKES AND MODELS USED IN THIS STUDY}\vspace{4pt}
\begin{tabular}{|c|c|}
		\hline
 		\textbf{Make name} & \textbf{Model name} \\
 		\hline
		Ferrari & Purosangue \\
		\hline
		Kia & EV9 \\
		\hline
		Lamborghini & Revuelto \\
		\hline
		Mazda & EZ6 \\
		\hline
		Mitsubishi & Xforce \\
		\hline
		Nissan & Ariya \\
		\hline
		Rolls Royce & Spectre \\
		\hline
		Toyota & Supra GRMN \\
		\hline
		Volkswagen & ID.Buzz \\
		\hline
		Volvo & EX30 \\
		\hline
\end{tabular}
\label{tabCMM}
\end{table}

\subsection{Experimental Setup}

\begin{figure*}[!t]
\centering
\includegraphics[width=0.95\linewidth]{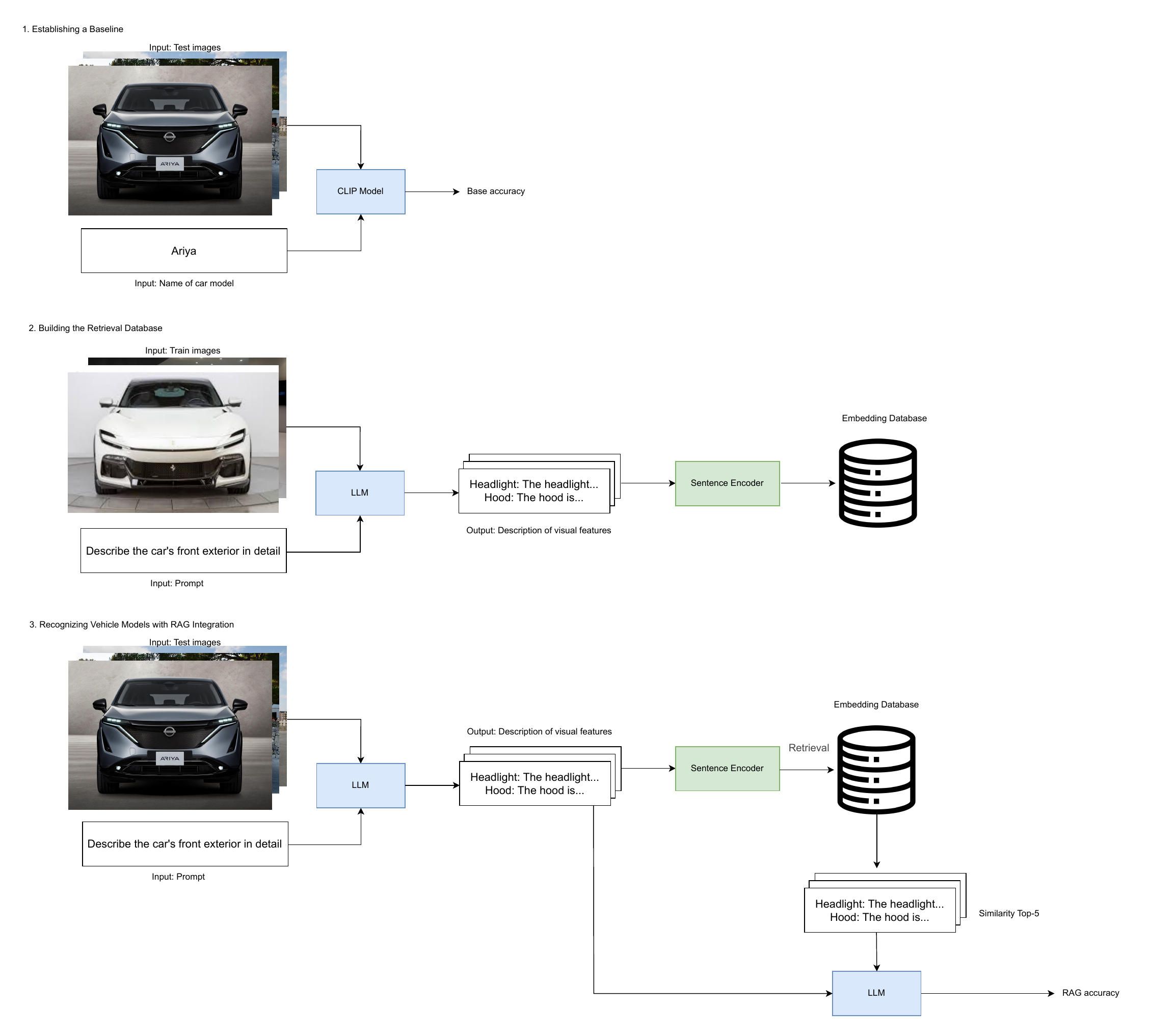}
\caption{Experimental Framework for Car Model Recognition: CLIP vs. RAG Approach}
\label{fig:architecture}
\end{figure*}

Fig. \ref{fig:architecture} illustrates the full experimental architecture, including the RAG integration within the recognition pipeline. We employ CLIP as the text encoder to obtain both image and text embeddings, while the large language model is Gemma 3 \cite{gemma3}, an open-source model built upon Gemini 2.0\cite{gemini}. The 12B parameter version of Gemma~3 is used to generate fine-grained exterior feature descriptions and to reason over retrieved context. The retrieval database is populated with LLM-generated descriptions of training samples, which are encoded and stored in a FAISS\cite{faiss} vector index for cosine-similarity search. All experiments are conducted on a single NVIDIA GeForce RTX 5090 GPU, and the same pipeline is applied for both the baseline and RAG-enhanced evaluations.

\subsection{Experimental Procedure}
The experimental procedure can be divided into three steps: 
\subsubsection{Establishing a Baseline}
To establish a baseline, we employ CLIP Vit-B/32 for zero-shot car model recognition. The ten car models listed in Table \ref{tabCMM} are treated as textual labels. Each of the 100 test images is sequentially input to the CLIP model, which computes the cosine similarity between the image and each textual label. The label with the highest similarity is taken as the predicted class, and the overall prediction accuracy is then calculated.

\subsubsection{Building the Retrieval Database}
After establishing the baseline, we input the 10 training images to the LLM to generate front-end feature descriptions. These descriptions, along with their corresponding labels, are formatted as JSON and encoded into text embeddings using the CLIP ViT-B/32 text encoder, then stored in a FAISS\cite{faiss} vector database.

\subsubsection{Recognizing Vehicle Models with RAG Integration}
We re-input the 100 test images to LLM to generate feature descriptions, encode them with CLIP ViT-B/32, and retrieve the top-k similar entries from the database via cosine similarity, where k represents the number of retrieved documents and is set to 5 by default. The generated description and retrieved entries are combined into a new prompt, which is then input to the LLM for prediction. The LLM compares the generated description with each retrieved entry and selects the most similar one as the predicted car model category.

\subsubsection{Calculating Model Performance}
After obtaining predictions for all test images, we visualize the results as a confusion matrix and compute standard metrics, including accuracy, precision, and recall. These indicators collectively assess both overall and class-level performance in the recognition task.

\subsection{Performance Evaluation}
As shown in Table \ref{tab:overall-results}, the metrics show the distinctions in overall performance for car model recognition between our RAG approach and the CLIP model. The CLIP baseline achieved 21\% accuracy, while our RAG-based method reached 37\% accuracy, representing a notable improvement of 16 percentage points compared to the CLIP baseline, with similar gains observed in precision and recall metrics. This improvement shows our RAG method demonstrates better adaptability to new car models without requiring extensive image datasets for training. Since the dataset contains balanced samples across all categories, the recall and precision values are identical 
to the accuracy scores. We report all three metrics for each car model to provide a complete evaluation (Table \ref{tab:each-results-clip}, \ref{tab:each-results-rag}).

\begin{table}[!t]
    \centering
    \caption{OVERALL PERFORMANCE METRICS FOR RAG AND CLIP METHODS}
    \begin{tabular}{|c|c|c|c|}
    \hline
        Model & Accuracy & Recall & Precision \\
        \hline
        CLIP & 0.21 & 0.21 & 0.22 \\
        \hline
        RAG-based LLM (Ours) & 0.37 & 0.37 & 0.40 \\
        \hline
    \end{tabular}
    \label{tab:overall-results}
\end{table}

\begin{figure}[!t]
    \centering
    \begin{subfigure}[b]{0.45\textwidth}
        \includegraphics[width=\textwidth]{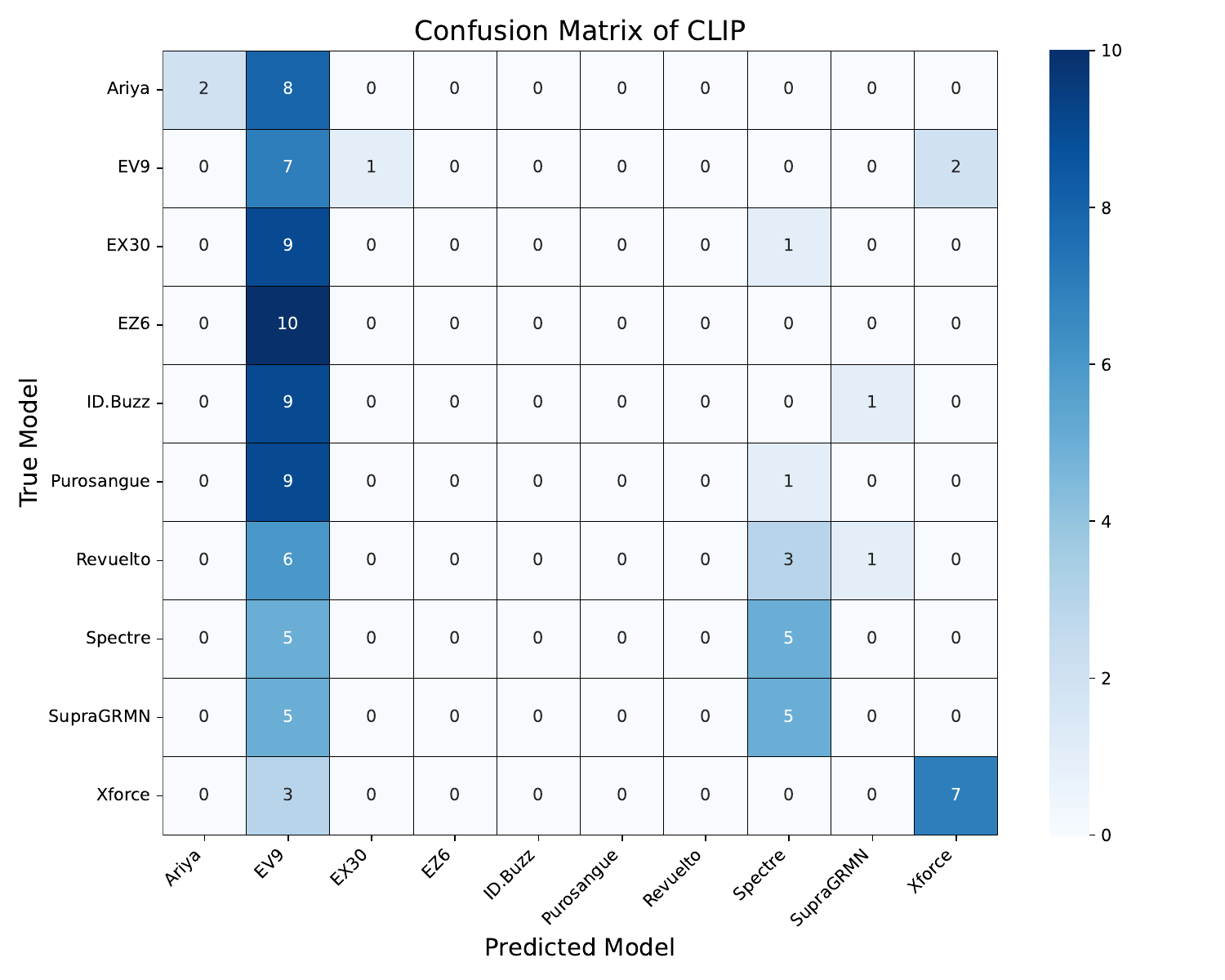}
        \caption{CLIP Model}
        \label{fig:sub-a}
    \end{subfigure}
    \hfill 
    \begin{subfigure}[b]{0.45\textwidth}
        \includegraphics[width=\textwidth]{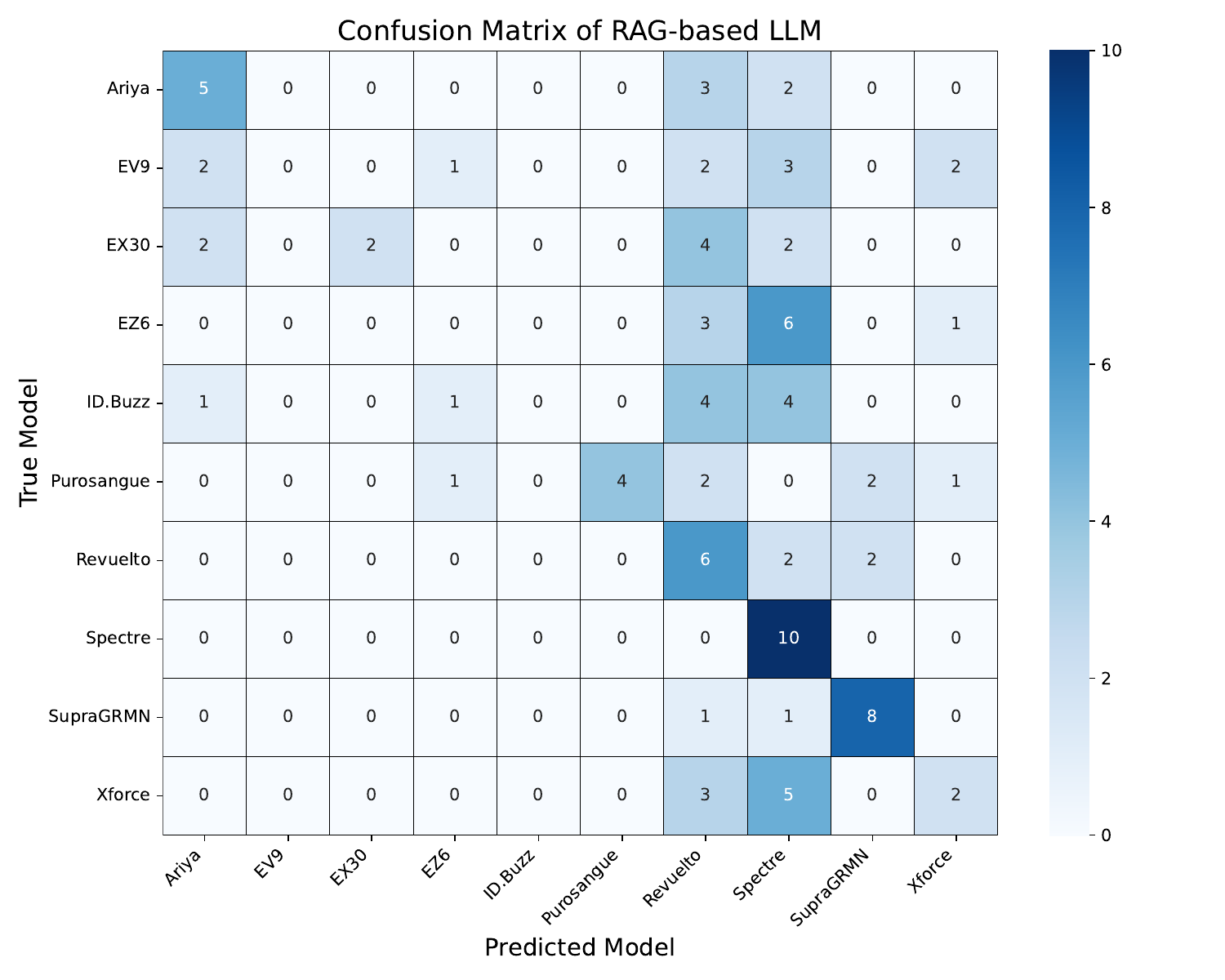}
        \caption{RAG-based LLM (Ours)}
        \label{fig:sub-b}
    \end{subfigure}
    \caption{Confusion Matrices for Car Model Classification: CLIP vs. RAG Approach}
    \label{fig:confusion-matrix}
\end{figure}

\begin{table}[!t]
    \centering
    \caption{DETAILED PERFORMANCE METRICS FOR EACH CAR MODEL IN CLIP}
    \begin{tabular}{|c|c|c|c|}
        \hline
        Car Model & Accuracy & Recall & Precision \\
        \hline
        Ariya & 0.20 & 0.20 & 1.00 \\
        \hline
        EV9 & 0.70 & 0.70 & 0.10 \\
        \hline
        EX30 & 0.00 & 0.00 & 0.00 \\
        \hline
        EZ6 & 0.00 & 0.00 & 0.00 \\
        \hline
        ID.Buzz & 0.00 & 0.00 & 0.00 \\
        \hline
        Purosangue & 0.00 & 0.00 & 0.00 \\
        \hline
        Revuelto & 0.00 & 0.00 & 0.00 \\
        \hline
        Spectre & 0.00 & 0.50 & 0.33 \\
        \hline
        SupraGRMN & 0.50 & 0.00 & 0.00 \\
        \hline
        Xforce & 0.70 & 0.70 & 0.78 \\
        \hline
    \end{tabular}
    \label{tab:each-results-clip}
\end{table}

\begin{table}[!t]
    \centering
    \caption{DETAILED PERFORMANCE METRICS FOR EACH CAR MODEL IN RAG-BASED LLM}
    \begin{tabular}{|c|c|c|c|}
        \hline
        Car Model & Accuracy & Recall & Precision \\
        \hline
        Ariya & 0.50 & 0.50 & 0.50 \\
        \hline
        EV9 & 0.00 & 0.00 & 0.00 \\
        \hline
        EX30 & 0.20 & 0.20 & 1.00 \\
        \hline
        EZ6 & 0.00 & 0.00 & 0.00 \\
        \hline
        ID.Buzz & 0.00 & 0.00 & 0.00 \\
        \hline
        Purosangue & 0.40 & 0.40 & 1.00 \\
        \hline
        Revuelto & 0.60 & 0.60 & 0.21 \\
        \hline
        Spectre & 1.00 & 1.00 & 0.29 \\
        \hline
        SupraGRMN & 0.80 & 0.80 & 0.67 \\
        \hline
        Xforce & 0.20 & 0.20 & 0.33 \\
        \hline
    \end{tabular}
    \label{tab:each-results-rag}
\end{table}

Fig. \ref{fig:confusion-matrix} shows the confusion matrices for both the CLIP (Fig. \ref{fig:sub-a}) and RAG (Fig. \ref{fig:sub-b}) approaches, providing detailed insights into the classification accuracy for each car model category. The experimental results show that our method not only outperforms CLIP in overall metrics but also achieves more balanced performance across different car model categories.

\subsection{Varying K Values \label{sec:diffK}}
\begin{figure}[h]
    \centering
    \includegraphics[width=1\linewidth]{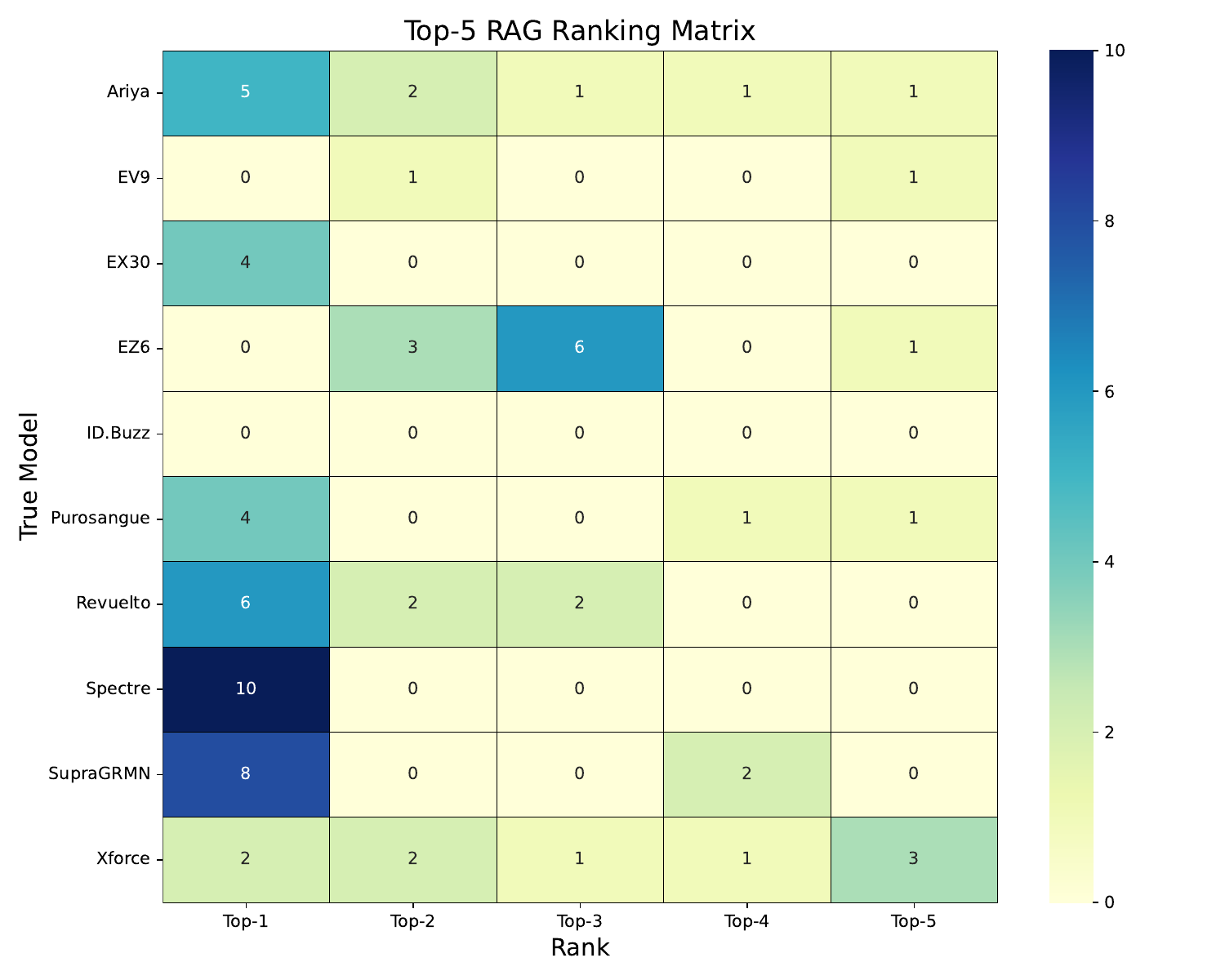}
    \caption{Retrieval Rank Distribution of True Models in Top-5 RAG Results}
    \label{fig:Top-5}
\end{figure}

Initially, we chose a midpoint value of 5 for the K-value out of the 10 total car models. The experimental results showed that the accuracy already increased by 16\%. We observed from Fig. \ref{fig:Top-5} that most of the correct retrievals were concentrated within the top-3. Therefore, we modified the selection of the K-value, and the test results are shown in Table \ref{tab:top-k}. Although the difference was not significant, it still suggests that too many options can lead to incorrect predictions by the LLM during the final decision-making, even if the retrieval is correct. In this experiment, a Top-1 K-value achieved the best model recognition performance, improving by 18\% compared to the CLIP baseline.

\begin{table}[h]
    \centering
    \caption{ACCURACY OF DIFFERENT K-VALUE}
    \begin{tabular}{|c|c|}
        \hline
        K-value & Accuracy\\
        \hline
        CLIP Model & 0.21 \\
        \hline
        Top-1 & 0.39 \\
        \hline
        Top-3 & 0.38 \\
        \hline
        Top-5 & 0.37 \\
        \hline
        Top-7 & 0.33 \\
        \hline
    \end{tabular}
    \label{tab:top-k}
\end{table}

\subsection{Investigating Prediction Errors in Image Comparison}
\begin{figure}[h]
  \centering
  \begin{subfigure}{0.45\linewidth}
    \centering
    \includegraphics[width=\linewidth]{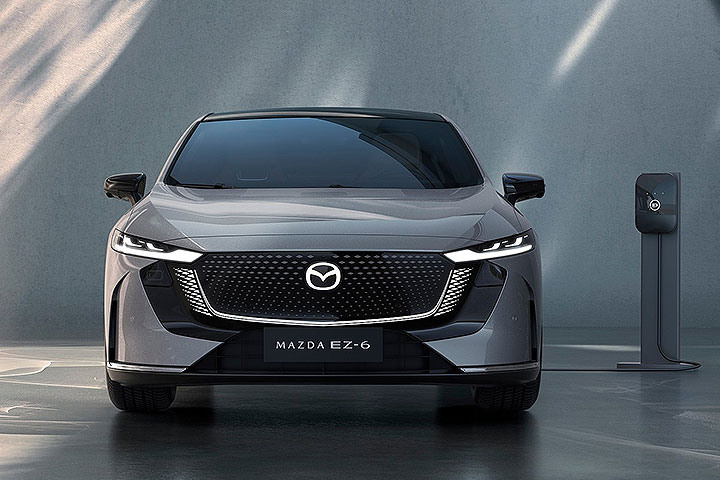}
    \caption{Mazda EZ6}
    \label{fig:ez6}
  \end{subfigure}
  \hfill
  \begin{subfigure}{0.45\linewidth}
    \centering
    \includegraphics[width=\linewidth]{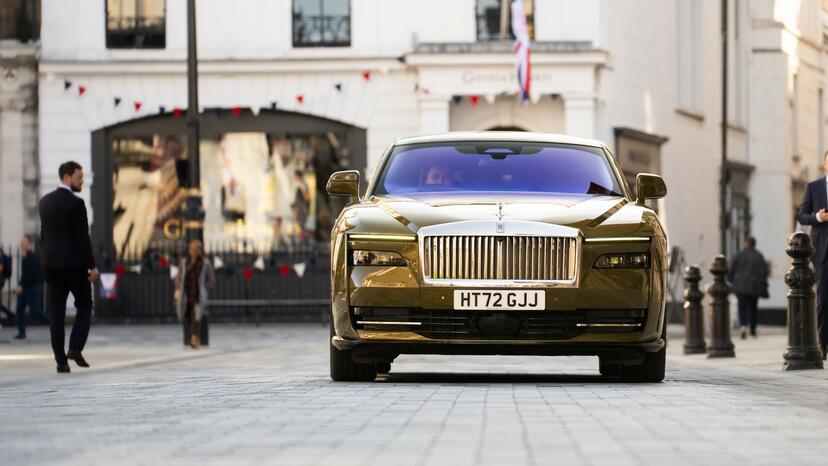}
    \caption{Rolls Royce Spectre}
    \label{fig:spectre}
  \end{subfigure}
  \caption{EZ6 and Spectre: A Comparative View}
  \label{fig:ez6andspectre}
\end{figure}

We analyzed the vehicles with the highest prediction errors from Fig. \ref{fig:sub-b}, namely the EZ6 and the Spectre. As shown in Figures \ref{fig:ez6} and \ref{fig:spectre}, which are test images of the two cars, their appearances are not highly similar. However, in the textual descriptions, the model tended to emphasize overall geometric proportions and shapes while overlooking brand-specific details (such as the chrome trim design and air intake grille texture). For example, descriptions mentioning horizontal and narrow headlights, a simple and horizontally extended grille outline, and smooth hood lines without distinct creases caused the descriptions for the EZ6 and Spectre to be highly similar, leading to confusion. 

This issue may stem from the generic nature of the LLM. The LLM we used is a general-purpose model not specifically designed for image understanding and detailed feature generation. It has limited capabilities in processing fine image details and variations in light and shadow, and it tends to generate generalized descriptions that lack precision.

\subsection{Limitations}
Our RAG-enhanced method, while demonstrating improved performance, has the following four key limitations:
\begin{itemize}
    \item \textbf{Dataset Limitations:} Our evaluation was conducted on a balanced dataset, which may not reflect the imbalanced distribution commonly encountered in real-world scenarios.
    \item \textbf{Increased Inference Time:} The retrieval process introduces additional computational overhead, resulting in longer prediction times compared to direct CLIP inference.
    \item  \textbf{Dependency on Description Quality:} The method's effectiveness is heavily dependent on the accuracy and quality of LLM-generated descriptions, which may introduce errors or inconsistencies.
    \item \textbf{Recognition Performance Constraints:} For common car models, text-based retrieval accuracy still falls short of pure computer vision methods. Our approach is more suitable as a rapid auxiliary tool for zero-shot recognition of new or rare car categories rather than a replacement for established visual recognition methods.
\end{itemize}
However, it is worth noting that the limitation regarding description quality dependency shows promising prospects for resolution. With the rapid advancement of multimodal LLMs and their improved capability in processing visual features, this constraint is expected to be gradually addressed. Future developments in LLM technology, including better fine-tuning approaches and more sophisticated multimodal architectures, are likely to significantly enhance the reliability of generated descriptions.

\section{Conclusion}
\label{sec:conclude}
This paper presents a text-retrieval-based car model recognition method that establishes a retrieval database through textual descriptions of automotive visual features and performs predictions based on these textual data. Unlike traditional image recognition approaches that rely on extensive training datasets, our method demonstrates significant improvements in recognizing new car models compared to the CLIP baseline. With appropriate selection of top-k values for retrieval, our approach can achieve performance improvements of nearly 20\%. These results highlight the potential and advantages of RAG-based LLMs in zero-shot image recognition tasks, not only reducing LLM hallucinations but also providing better adaptability to evolving car model updates.

Looking forward, this work opens several promising research avenues, including optimization of retrieval efficiency, development of hybrid visual-textual approaches, and expansion to broader fine-grained recognition applications as multimodal LLM technology continues to advance.

\bibliographystyle{IEEEtran}

\bibliography{bib}



\end{document}